\documentclass[runningheads]{llncs}
\usepackage{graphicx}

\usepackage{tikz}
\usepackage{comment}
\usepackage{amsmath,amssymb} 
\usepackage{color}

\usepackage{graphicx}
\usepackage{amsmath,amssymb} 
\usepackage{times}
\usepackage{epsfig}
\usepackage{graphicx}
\usepackage{amsmath}
\usepackage{amssymb}
\usepackage{colortbl}
\usepackage{relsize}
\usepackage{multirow}
\usepackage{float}

\begin{document}
%
\title{Recursive Deformable Image Registration Network with Mutual Attention}
%
\author{Jian-Qing Zheng\inst{1,2} \and
Ziyang Wang\inst{3} \and
Baoru Huang\inst{4}
\and Tonia Vincent\inst{1}
\and Ngee Han Lim\inst{1} \and Bart{\l}omiej W. Papie{\.z}\inst{2,5}}
%
\authorrunning{J.-Q. Zheng et al.}
%
\institute{The Kennedy Institute of Rheumatology, University of Oxford, U.K.
\and
Big Data Institute, University of Oxford, U.K.
\and
Department of Computer Science, University of Oxford, U.K.
\and
Department of Surgery and Cancer, Imperial College London
\and
Nuffield Department of Population Health, University of Oxford, UK
\email{\{jianqing.zheng@kennedy,bartlomiej.papiez@bdi\}.ox.ac.uk}}


%

\maketitle              
\begin{abstract}
Deformable image registration, estimating the spatial transformation between different images, is an important task in medical imaging.
Many previous studies have used learning-based methods for multi-stage registration to perform 3D image registration to improve performance. 
The performance of the multi-stage approach, however, is limited by the size of the receptive field where complex motion does not occur at a single spatial scale. 
We propose a new registration network combining recursive network architecture and mutual attention mechanism to overcome these limitations. 
Compared with the state-of-the-art deep learning methods, our network based on the recursive structure achieves the highest accuracy in lung Computed Tomography (CT) data set (Dice score of 92\% and average surface distance of 3.8mm for lungs) and one of the most accurate results in abdominal CT data set with 9 organs of various sizes (Dice score of 55\% and average surface distance of 7.8mm).
We also showed that adding 3 recursive networks is sufficient to achieve the state-of-the-art results without a significant increase in the inference time.

\end{abstract}
\section{Introduction}
Deformable image registration (DIR) is an essential computer vision task which has been widely studied \cite{sotiras2013deformable}.
In medical imaging, DIR enables the estimation of the non-linear correspondence between different acquisitions over time to monitor progress of treatment, or between different types of scanners (e.g. multi-modal image fusion) to provide complementary disease information. 
The classical registration algorithms have been developed as continuous optimization \cite{avants2008symmetric,rueckert1999nonrigid,thirion1998image}, 
or discrete optimization problems \cite{heinrich2013}. 
Their computational performance, however, is limited due to highly dimensional, non-convex problem, and low capability to capture complex, global and local deformations \cite{schnabel2016}. 
Recently, researchers have shifted interest to deep-learning-based unsupervised learning methods in deformable image registration, because data-driven methods benefit significantly from a large amount of given paired/unpaired images compared with classical methods \cite{jia2021learning,de2019deep,aggarwal2018modl}. A fast learning-based approach, VoxelMorph, is presented in \cite{balakrishnan2019voxelmorph}, where convolutional neural networks (CNN) and spatial transformer layers \cite{jaderberg2015spatial} are used to register two images by regressing directly dense displacement field. 
Other deep learning approaches investigated different representations of the transformation e.g. diffeomorphism \cite{mok2020fast}, which preserve the topology of the transformation. 
The direct regression of the spatial transformation via neural networks however, only gives one prediction on registration without any progressive refinement.

Multi-stage architecture is one of the solutions that is beneficial to CNN \cite{de2019deep,hu2018weakly,zhao2019unsupervised}. A weakly supervised multi-model registration method \cite{hu2018weakly}, utilizing an end-to-end convolution based network, aims to predict displacement fields to align multiple labelled corresponding structures for individual image pairs.
Alternatively, an end-to-end multi-stage networks \cite{zhao2019recursive} are proposed for a deep recursive cascade architecture that allows unlimited number of cascades that can be built on the backbone networks. All of these multi-stage cascaded network structures, however, are still potentially suffering from the limited capture range of the receptive field.

The attention mechanism \cite{vaswani2017attention} addresses the limited receptive field of CNNs and has been widely utilized in transformer networks.
Optimal correspondence matching was studied in \cite{li2021revisiting} for a stereo matching task, where self-attention-based transformer is proposed to relax the limitation of a fixed disparity range. Local feature matching can also benefit from self and cross attention, because transformer networks are proved to obtain feature descriptors that are conditioned on both images \cite{sun2021loftr}.
The attention-based mechanism was applied to registration \cite{zhang2021learning,song2021cross,chen2021transmorph} previously, however is computationally expensive, and thus has not been explored in recursive deformable image registration.

In this paper, we propose a Recursive Mutual Attention Network (RMAn), combining the Mutual Attention (MA) based module with a recursive architecture to increase the size of the receptive field. 
The recursive architecture provides the progressive refinement to 3D image deformable registration so that MA module can expand the global receptive field on a pair of low-resolution feature maps without extra cost of computation. Our contributions in this paper are as follows.
\begin{enumerate}
    \item A Mutual Attention based Recursive Network (RMAn) is proposed for deformable image registration, combining the mutual attention \cite{vaswani2017attention} into recursive networks \cite{zhao2019recursive};
    \item The proposed network achieves the best performance against the state-of-the-art  deep learning network structures respectively in lung Computed Tomography (CT) data set (Dice similarity coefficient of 92\% and average surface distance for of 3.8mm lung) and comparable performance in abdomen (9 organs) CT data set.
\end{enumerate}

\section{Methods}
\label{sec:methods}
\begin{figure}[t]
\begin{center}
\includegraphics[width=1.\linewidth]{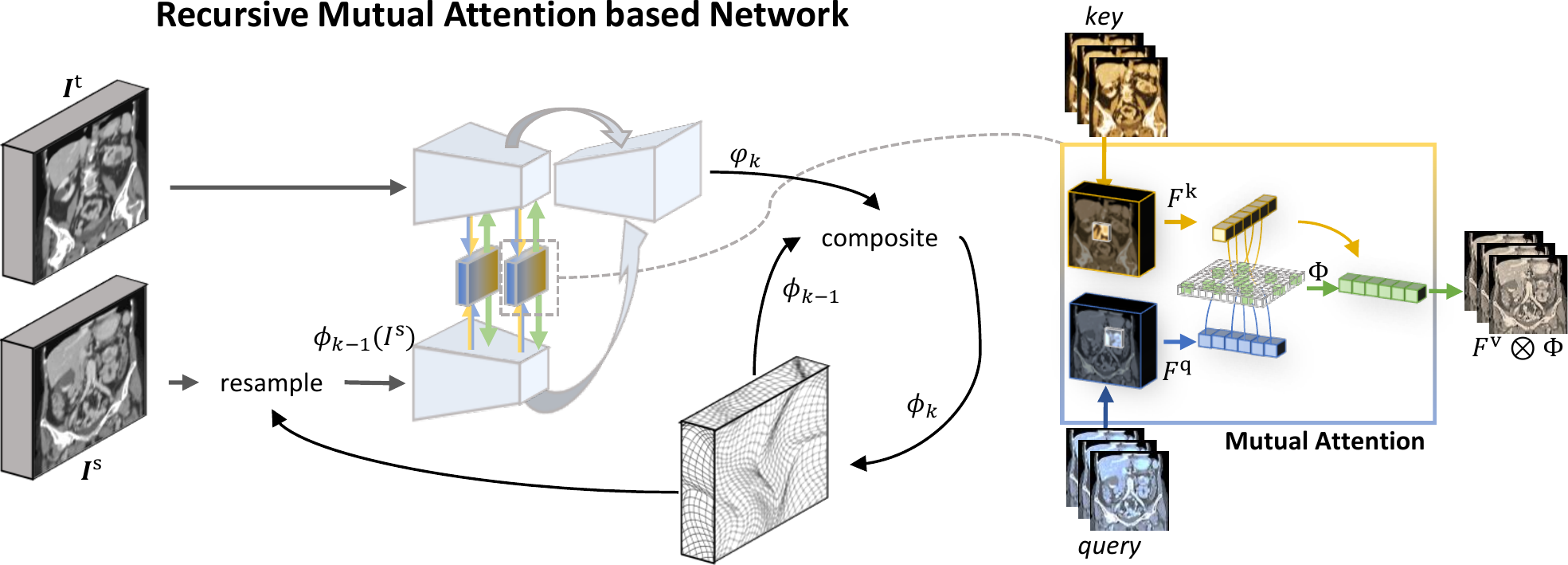}
\end{center}
\caption{Proposed framework of Recursive Mutual Attention based Network, including a Siamese Encoder-Decoder structure with Mutual Attention interconnected, and the network structure detailed in Fig.~\ref{fig:modules}, where $k\in[1,K]\cap\mathbb{Z}$ denotes the recursive index and $K\in\mathbb{Z}_+$ denotes the total recurrent number.}
\label{fig:networks}
\end{figure}
\begin{figure}[t]
\begin{center}
\includegraphics[width=1.\linewidth]{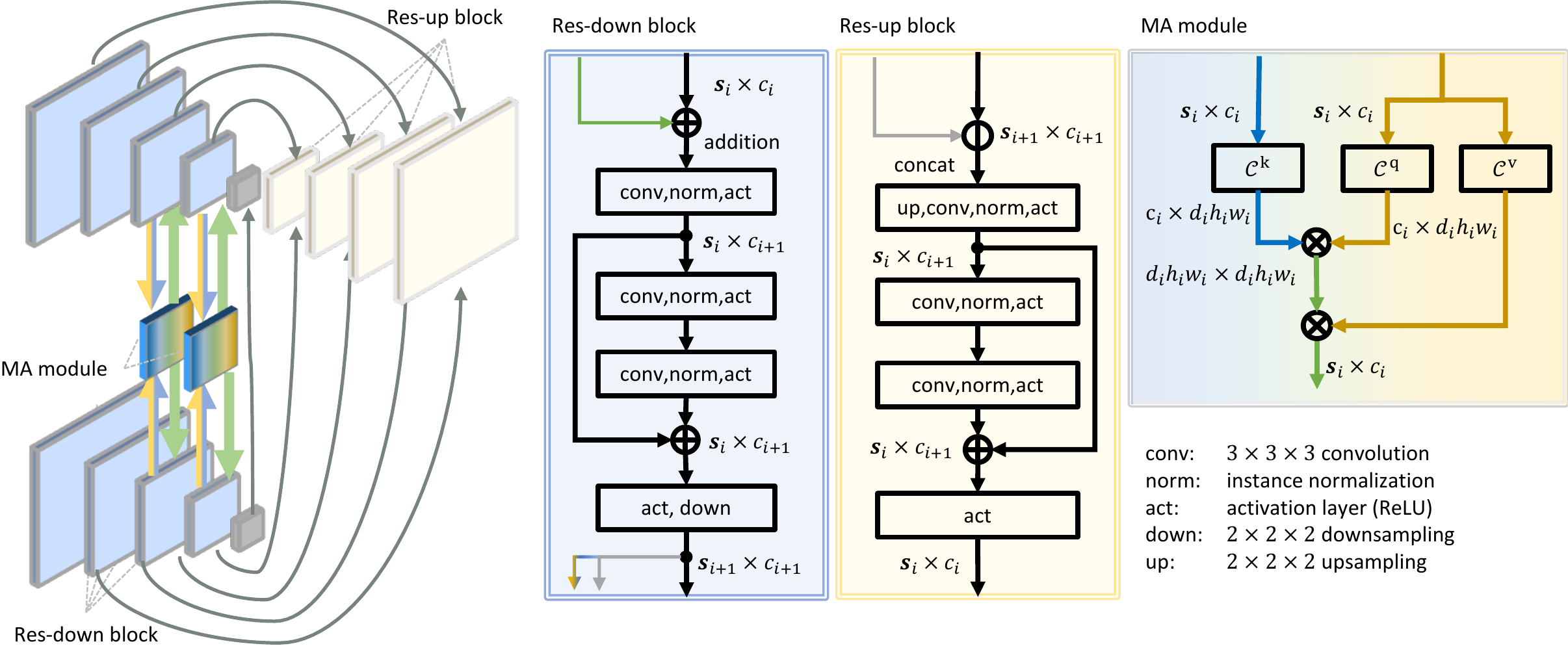}
\end{center}
\caption{The subnetwork in Fig.~\ref{fig:networks} including three main components, a Siamese Encoder consists of four pairs of Residual Downsampling (Res-down) blocks, Residual Upsampling (Res-up) block, and two Mutual Attention (MA) modules.}
\label{fig:modules}
\end{figure}
\subsection{Image Registration}
\label{sec:image_reg}
Image registration can be defined as estimation of the spatial transformation ${\phi}:\mathbb{R}^{n}\to\mathbb{R}^{n}$, represented by the corresponding parameters or a series of displacements denoted by $\phi[\textbf{\textit{x}}]\in \mathbb{R}^{d}$ at the spatial position $\textbf{\textit{x}}\in\mathbb{Z}^{d}$ of a target image $\textbf{\textit{I}}^{\rm t}\in\mathbb{R}^{{n}}$ from a source image  $\textbf{\textit{I}}^{\rm s}\in\mathbb{R}^{n}$, where $n$ is the size of a 3D image defined as $n={H \times W\times T}$, and $d, T, H, W$ denoting the image dimension, thickness, height, and width, respectively.
Originally, image registration was solved as an optimization problem by minimization of a dissimilarity metric $\mathcal{D}$ and a regularization term $\mathcal{S}$:
\begin{equation}
\label{equ:opt_phi}
\hat{\phi}=\underset{\phi}{\mathrm{argmin}}{\big(\mathcal{D}(\phi(\textbf{\textit{I}}^{\rm s}),~\textbf{\textit{I}}^{\rm t})+\lambda\mathcal{S}(\phi,\textbf{\textit{I}}^{\rm t})\big)}
\end{equation}
where $\hat{\phi}$ denotes the estimated spatial transformation,
$\lambda$ denotes the weight of the regularization. 
More recently, the registration is performed 
via CNN $\mathcal{R}$ directly regressing the spatial transformation e.g. using the Dense Displacement Field (DDF)  \cite{balakrishnan2018unsupervised,mok2020fast}:
\begin{equation}
\phi=\mathcal{R}(\textbf{\textit{I}}^{\rm s},\textbf{\textit{I}}^{\rm t};w)
\end{equation}
with the training process based on minimizing the loss function (e.g. given in Eq.~\eqref{equ:opt_phi}) with the trainable weights $w$ 
($w$ is omitted in the following part of the paper to simplify the formula). 
However the direct regression of spatial transformations via convolution neural networks could suffer due to limited capture range of the receptive field of convolutional layers when dealing with large motion.

\subsection{Recursive Registration Networks}
\label{sec:rec_reg}
Inspired by \cite{zhao2019recursive}, we proposed a recursive network structure for coarse-to-fine registration of a pair of images as shown in Fig.~\ref{fig:networks}.
In coarse-to-fine approach, the residual transformation $\varphi_{k}$ between the target image $\textbf{\textit{I}}^{\rm t}$ and the warped source feature map based on previous level $k-1$ registration $\phi_{k-1}(\textbf{\textit{I}}^{\rm s})$ 
is estimated via $\mathcal{R}$ and accumulated via composition:
\begin{equation}
\label{equ:res_align}
\left\{
\begin{array}{cc}
    \phi_k=\phi_{k-1}\circ\varphi_{k}\\
    \varphi_{k}=\mathcal{R}(\phi_{k-1}(\textbf{\textit{I}}^{\rm s}),\textbf{\textit{I}}^{\rm t})\\
\end{array}
\right.
\end{equation}
where 
$\circ$ denotes the composition of two spatial transformations, and $\phi_{0}$ is initialized as the identity transform.
The subnetwork used in Fig.~\ref{fig:networks} including a weight-sharing two-branch Siamese encoder interconnected with a Mutual Attention module to extract and retrieve the common features, and a decoder to estimate the DDF $\varphi_k$, where each component of the network structure is shown in Fig.~\ref{fig:modules}, and where the convolution layers in each Res-down and Res-up blocks are set with the kernel size of 3, and atrous rate (1,1,3) following the theoretical optimization of receptive field size in \cite{zhou2020acnn}.

\subsection{Mutual Attention}
\label{sec:mutual_attention}
Similar to the idea from \cite{li2021revisiting,sun2021loftr,heinrich2019closing,zheng2020d}, Mutual Attention (MA) mechanism \cite{vaswani2017attention} is used in the RMAn to obtain the global receptive field and use so-called indicator matrices to quantify the relationship between each pair of pixels from two images, and the usage of multiple indicator matrices is called multi-head. 
The feature maps $\textbf{\textit{F}}^{\rm k},\textbf{\textit{F}}^{\rm q}\in\mathbb{R}^{c\times n}$ are extracted from two stream of the two images $\textit{\textbf{I}}^{\rm s},\textit{\textbf{I}}^{\rm t}$ via the encoder part as shown in Fig.~\ref{fig:networks}, where $c$ denotes the feature channel number.
Each element of $\textbf{\textit{F}}^{\rm k}$ (yellow arrow in Fig.~\ref{fig:networks} and Fig.~\ref{fig:modules}) as a key vector is retrieved in the query vectors via correlation from $\textbf{\textit{F}}^{\rm q}$ (blue arrow) in an indicator matrix $\Phi\in\mathbb{R}^{n\times n}$ which can be formulated as:
\begin{equation}
\label{equ:cross_attn}
\Phi={\rm softmax}(\mathcal{C}^{\rm q}({\textbf{\textit{F}}^{\rm k}})^{\top}\mathcal{C}^{\rm k}({\textbf{\textit{F}}^{\rm q}}))
\end{equation}
Then the vector from $\textbf{\textit{F}}^{\rm q}$ is passed through the corresponding linear mapping to the other stream via the $\Phi$:
\begin{equation}
\label{equ:cross_attn_pass}
\left\{
\begin{array}{llr}
{\textbf{\textit{F}}^{\rm v}}=\mathcal{C}^{\rm v}(\textbf{\textit{F}}^{\rm k})\\
{\textbf{\textit{F}}^{\rm k\to q}}=\textbf{\textit{F}}^{\rm v}\otimes\Phi
\end{array}
\right.
\end{equation}
where ${\textbf{\textit{F}}^{\rm k\to q}}$ denotes the feature maps passed from one stream to the other (green arrow in Fig.~\ref{fig:networks} and Fig.~\ref{fig:modules}), $\mathcal{C}^{\rm q}$, $\mathcal{C}^{\rm k}$ and $\mathcal{C}^{\rm v}$ denote the linear transformation for query, key and value feature vectors, respectively. Because the MA module is used bi-directly, the feature forwarded as both the key and query features are denoted as half blue half yellow arrows in Fig.~\ref{fig:networks} and Fig.~\ref{fig:modules}, and the corresponding green arrow always point to the branch of the query stream.

\section{Experiments}
\label{sec:experiments}
\subsection{Datasets}
We evaluated the proposed RMAn for unsupervised deformable registration problem using two publicly available data sets with the ground truth annotations for 9 organs in abdomen CT data set and lung volumes annotations in lung CT data set.

\subsubsection{Unpaired Abdomen CTs} are selected from~\cite{dalca2020learn2reg}. The ground truth segmentation of spleen, right kidney, left kidney, esophagus, liver, aorta, inferior vena cava, portal, splenic vein, and pancreas are annotated for all CT scans. The inter-subject registration of the abdominal CT scans is challenging due to large inter-subject variations and great variability in organ volume, from 10 milliliters (esophagus) to 1.6 liters (liver). Following the previously presented works, each volume is resized to $2\times2\times2mm^3$ in the pre-processing step. From totally 30 subjects, 23 and 7 are respectively used for training and testing, forming 506 and 42 different pairs of images.

\subsubsection{Unpaired Chest (Lung) CTs} are selected from~\cite{hering_alessa_2020_3835682}. The CT scans are all acquired at the same time point of the breathing cycle with a slice thickness of 1.00 mm and slice spacing of 0.70 mm. Pixel spacing in the X-Y plane varies from 0.63 to 0.77 mm with an average value of 0.70 mm. The ground truth annotations of lungs for all scans are provided. Following the previously presented works, each volume is resized to $1\times1\times1mm^3$ in the pre-processing step. We perform inter-subject registration from the total of 20 subjects, 12 and 8 are respectively used for training and testing, forming 132 and 56 different pairs of images.

\subsection{Training Details}
We normalize the input image into 0-1 range and augment the training data by randomly cropping input images during training. 
For the experiments on inter-subject registration of abdomen and lung CT, the models are first pre-trained for 50k iteration on synthetic DDF, with the loss function set as:
\begin{equation}
\mathcal{L}_{\rm syn}=\sum{{\|\phi-\tilde{\phi}\|}_2^2}+\lambda\sum{{\|\nabla\phi\|}_2^2}
\end{equation}
Then the models are trained on real data for 100k iterations with the loss function:
\begin{equation}
\mathcal{L}={\mathcal{D}(\textbf{\textit{I}}^{\rm t}-\phi(\textbf{\textit{I}}^{\rm s}))}+\lambda{{\|\nabla\phi\odot{{\rm e}^{-{\|\nabla\textbf{\textit{I}}^{\rm t}\|}_2^2}}\|}_2^2}
\end{equation}
where normalized cross correlation and mean squared error are used in abdomen and lung CT respectively for $\mathcal{D}$ following \cite{balakrishnan2019voxelmorph}.
The whole training takes one week, including the data transfer, pretraining and fine-tuning. With a training batch size of 3, The model was end-to-end trained with Adam optimizer with the initial learning rate set as 0.001.

\subsection{Implementation and Evaluation}
\subsubsection{Implementation:}
The code for inter-subject image registration tasks was developed based on the framework of \cite{balakrishnan2018unsupervised} in Python using Tensorflow and Keras. It was run on Nvidia Tesla P100-SXM2 GPU with 16GB memory, and Intel(R) Xeon(R) Gold 6126 CPU @ 2.60GHz. 
\subsubsection{Baselines:}
We compared RMAn with the relevant state-of-the-art networks. The Voxelmorph \cite{balakrishnan2019voxelmorph} is adopted as the representative state-of-the-art, deep learning method of direct regression (DR). 
The composite network combing CNN (Global-net) and U-net (Local-net) following to \cite{hu2018weakly}, recursive cascaded network (RCN) \cite{zhao2019recursive} were also adopted into the framework as the relevant baselines representing multi-stage (MS) networks, as well as D-net \cite{zheng2020d} was adopted for DIR based on the MA mechanism. 
\subsubsection{Evaluation Criterion:}
Following \cite{de2019deep}, we calculated the Dice Coefficient Similarity (DSC), Hausdorff Distance (HD), and Average Surface Distance (ASD) on annotated organs for the performance evaluation of nine organs in abdomen CT and one organ (lung) in chest CT. We additionally calculated the negative number of Jacobian determinant in tissues' region (detJ) for rationality evaluation on prediction. 
The model size and running time for comparison with the previous methods on inter-subject registration of lung and abdomen are shown in Tab.~\ref{tab:result_abdomen_lung}. 

\section{Results}

\begin{table}[h!]
\caption{Average of Dice Similarity Coefficient (DSC), Average Surface Distance (ASD), Hausdorff Distance (HD) and negative number of Jacobian determinant in tissues' region (detJ) for unsupervised inter-subject registration of abdomen and chest CT using the Voxelmorph (VM1) \cite{balakrishnan2019voxelmorph} and its enhanced version with double number of feature channels (VM2), D-net \cite{zheng2020d} adopted for deformable registration, convolution networks cascaded with U-net (Cn+Un) \cite{hu2018weakly}, 5-recursive cascaded network based on the structure of the Voxelmorph (RCn) \cite{zhao2019recursive}, and our RMAn network, with different registration (reg.) types and varying Parameter Number (\#Par), and Time cost per Pair of Images (TPI).}
\label{tab:result_abdomen_lung}
\begin{center}
\centering
\begin{tabular}{ |c|c|cccc|cccc|cc| }
\hline
\multirow{3}{*}{model} &\multirow{2}{*}{reg. }&\multicolumn{4}{c|}{abdomen (9 organs)}&\multicolumn{4}{c|}{chest (lung)}&\multicolumn{2}{c|}{efficiency}\\
&\multirow{2}{*}{type}
&\cellcolor[RGB]{255,170,170}DSC$\uparrow$ &\cellcolor[RGB]{153,204,255}HD$\downarrow$  &\cellcolor[RGB]{153,204,255}ASD$\downarrow$
&\cellcolor[RGB]{153,204,255} det{J}$\downarrow$
& \cellcolor[RGB]{255,170,170}DSC$\uparrow$ &\cellcolor[RGB]{153,204,255}HD$\downarrow$  &\cellcolor[RGB]{153,204,255}ASD$\downarrow$
&\cellcolor[RGB]{153,204,255} det{J}$\downarrow$
&\cellcolor[RGB]{153,204,255}\#Par$\downarrow$&\cellcolor[RGB]{153,204,255}TPI$\downarrow$
\\
&&\cellcolor[RGB]{255,170,170}(\%)&\cellcolor[RGB]{153,204,255}(mm)&\cellcolor[RGB]{153,204,255}(mm)&\cellcolor[RGB]{153,204,255}(e{3})&\cellcolor[RGB]{255,170,170}(\%)&\cellcolor[RGB]{153,204,255}(mm)&\cellcolor[RGB]{153,204,255}(mm)&\cellcolor[RGB]{153,204,255}(e{3})
&\cellcolor[RGB]{153,204,255}(e6)&\cellcolor[RGB]{153,204,255}(sec)
\\
\hline
\hline
Initial&--
&30.9&49.5&16.04&-- 
&61.9&41.6&15.86&--
& -- &--
\\
VM1 &DR
&44.7&\textbf{43.8}&9.24&2.23
&84.0&32.9&6.38&5.94
& 0.36& 0.23
\\
VM2 &DR
&51.9&45.0&8.40&4.03
&88.8&32.0&5.02&15.58
& 1.42& 0.25
\\
Dnet&MA
&47.4&47.6&8.72&5.28
&88.3&33.2&5.01&10.38
&0.40   &0.41
\\
Cn+Un &MS
&53.6&44.6&7.84&4.13
&91.1&\textbf{29.7}& 3.84&4.23
&2.11&0.36
\\ 
RCn &MS
&\textbf{55.6}&44.9&7.79&2.91
&89.8&33.1&4.68&5.68
&0.36 &0.44
\\  
\rowcolor[RGB]{230,230,230} RMAn&MS+MA
&55.2&45.1&\textbf{7.78}&{4.32}
&\textbf{92.0}&31.8&\textbf{3.83}&{4.53}
& 0.40& 0.67
\\
 \hline
\end{tabular}
\end{center}
\end{table}

\begin{figure}[h!]
\begin{center}
\includegraphics[width=.95\linewidth]{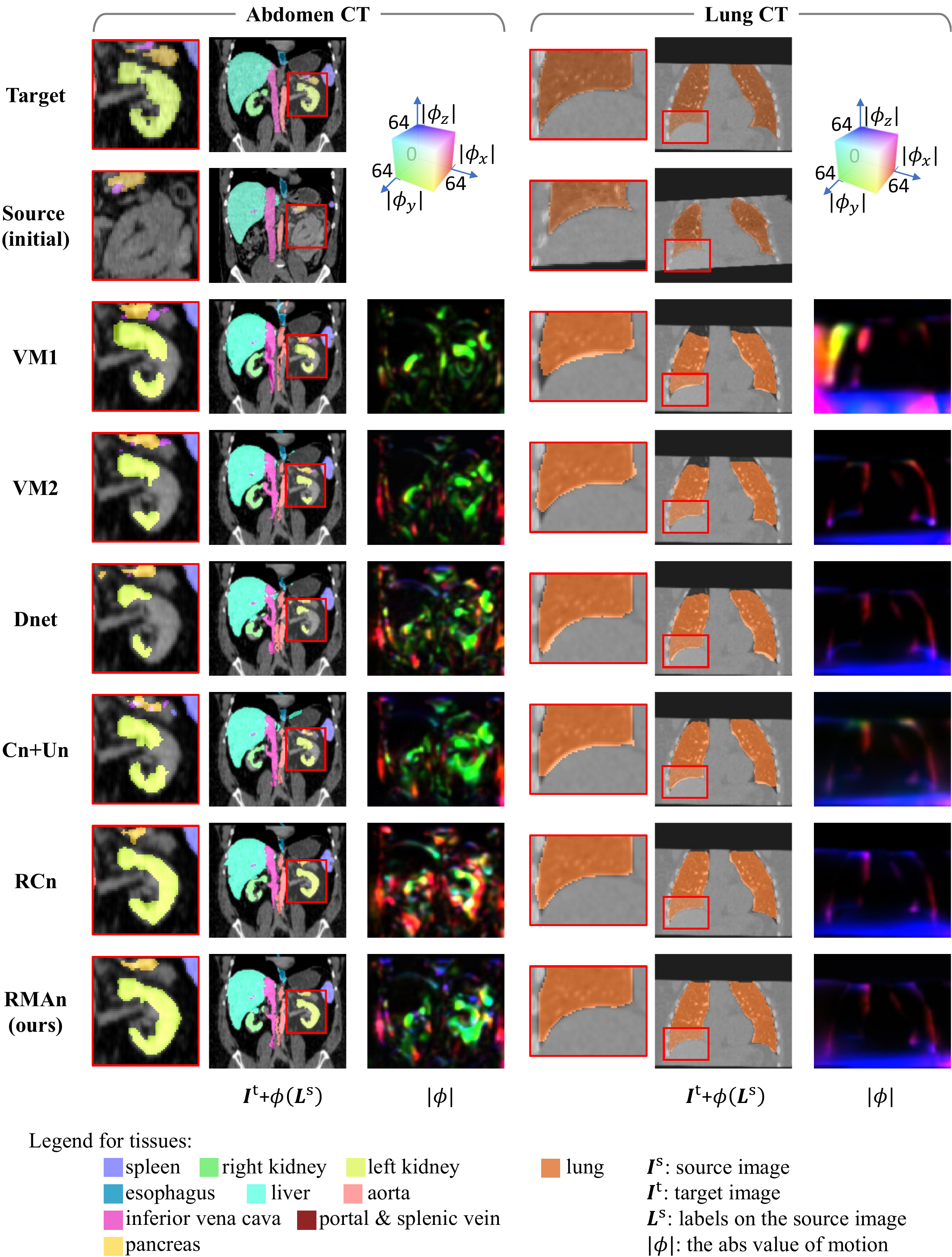}
\end{center}
\vspace{-1.5em}
\caption{Qualitative example in chest CT shows our network achieves plausible registration, with a significant improvement, especially at the edge area of the left kidney and the lung.}
\label{fig:qual_result}
\end{figure}

\begin{figure}[h!]
\begin{center}
\includegraphics[width=\linewidth]{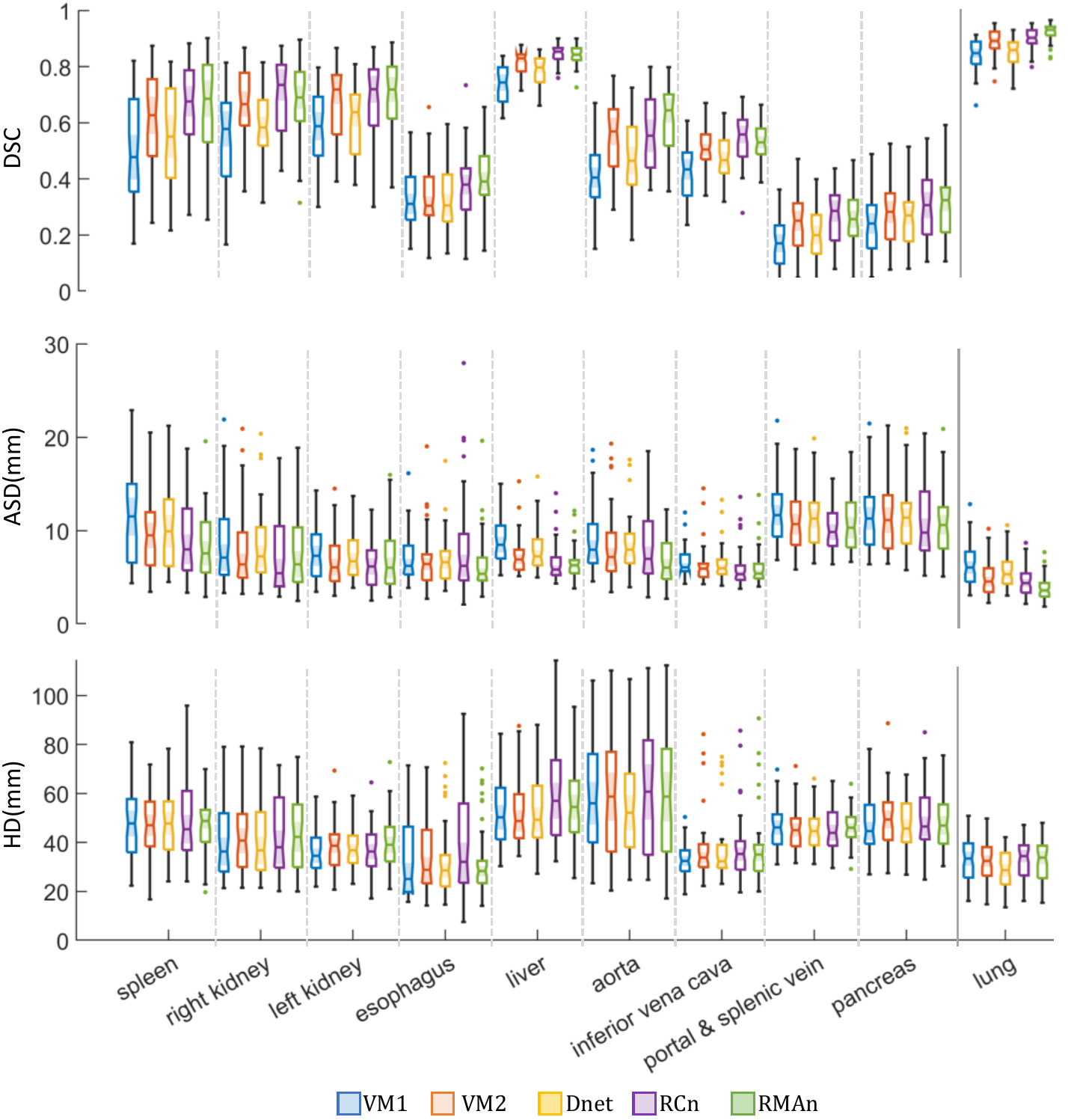}
\end{center}
\vspace{-1.5em}
\caption{RMANs achieve the best registration of the lung in chest CT scans as well as one of the best in the abdomen CT scans.}
\label{fig:quan_result}
\end{figure}

\begin{table}[h!]
\caption{Ablation study on recursive structure by inter-subject image registration of abdomen CT and lung CT using, with varying setting of recursive number (Rec.~No.) for training and testing.}
\label{tab:result_ablation}
\begin{center}
\centering
\begin{tabular}{ |c|cc|cccc|cccc|cc| }
\hline
\multirow{3}{*}{model} &\multicolumn{2}{c|}{Rec.~No.}&\multicolumn{4}{c|}{abdomen (9 organs)}&\multicolumn{4}{c|}{chest (lung)}&\multicolumn{2}{c|}{efficiency}\\
&\multirow{2}{*}{$K_{\rm train}$}&\multirow{2}{*}{$K_{\rm infer}$}
& \cellcolor[RGB]{255,170,170}DSC$\uparrow$ &\cellcolor[RGB]{153,204,255}HD$\downarrow$  &\cellcolor[RGB]{153,204,255}ASD$\downarrow$&\cellcolor[RGB]{153,204,255} det{J}$\downarrow$& \cellcolor[RGB]{255,170,170}DSC$\uparrow$ &\cellcolor[RGB]{153,204,255}HD$\downarrow$  &\cellcolor[RGB]{153,204,255}ASD$\downarrow$&\cellcolor[RGB]{153,204,255} det{J}$\downarrow$
&\cellcolor[RGB]{153,204,255}\#Par$\downarrow$&\cellcolor[RGB]{153,204,255}TPI$\downarrow$
\\
&&&\cellcolor[RGB]{255,170,170}(\%)&\cellcolor[RGB]{153,204,255}(mm)&\cellcolor[RGB]{153,204,255}(mm)&\cellcolor[RGB]{153,204,255}(e3)&\cellcolor[RGB]{255,170,170}(\%)&\cellcolor[RGB]{153,204,255}(mm)&\cellcolor[RGB]{153,204,255}(mm)&\cellcolor[RGB]{153,204,255}(e3)
&\cellcolor[RGB]{153,204,255}(e6)&\cellcolor[RGB]{153,204,255}(sec)\\
\hline
\hline
MAn&1&1
&47.4&47.6&8.72&5.28
&88.3&33.2&5.01&10.38
&0.40   &0.41
\\
RMAn &2&2
&52.2&45.5&8.35&{5.20}
&{91.2}&{\textbf{31.6}}&{4.16}&{6.74}
& 0.40& 0.64
\\    
RMAn &3&3
&54.4&44.9&7.91&{5.01}
&{91.4}&{32.6}&{{4.01}}&{5.36}
& 0.40& 0.65
\\
RMAn &3&5
&\textbf{55.2}&\textbf{45.1}&\textbf{7.78}&\textbf{4.32}
&{\textbf{92.0}}&31.8&{\textbf{3.83}}&\textbf{4.53}
&0.40 & 0.65
\\
 \hline
\end{tabular}
\end{center}
\end{table}

\subsubsection{Comparison with the state-of-the-art Networks:}
Our proposed RMAn is compared with other methods on unsupervised DIR of abdomen and chest CT using all 10 organs. With an intuitive qualitative results shown in Fig.~\ref{fig:qual_result}, RMAn achieves better performance on registration with an improvement in the area of lung boundaries (as depicted by the red box) and a plausible registration on the nine organs in the abdomen CT scans. The quantitative results shown in Fig.~\ref{fig:quan_result} illustrate that our RMAn achieves the best on lung and one of the best on the other nine abdominal organs. More numerical results are shown in Tab.~\ref{tab:result_abdomen_lung}, which demonstrates our network achieved comparable performance in this task with lower computational cost. 

\begin{figure}[t!]
\begin{center}
\includegraphics[width=\linewidth]{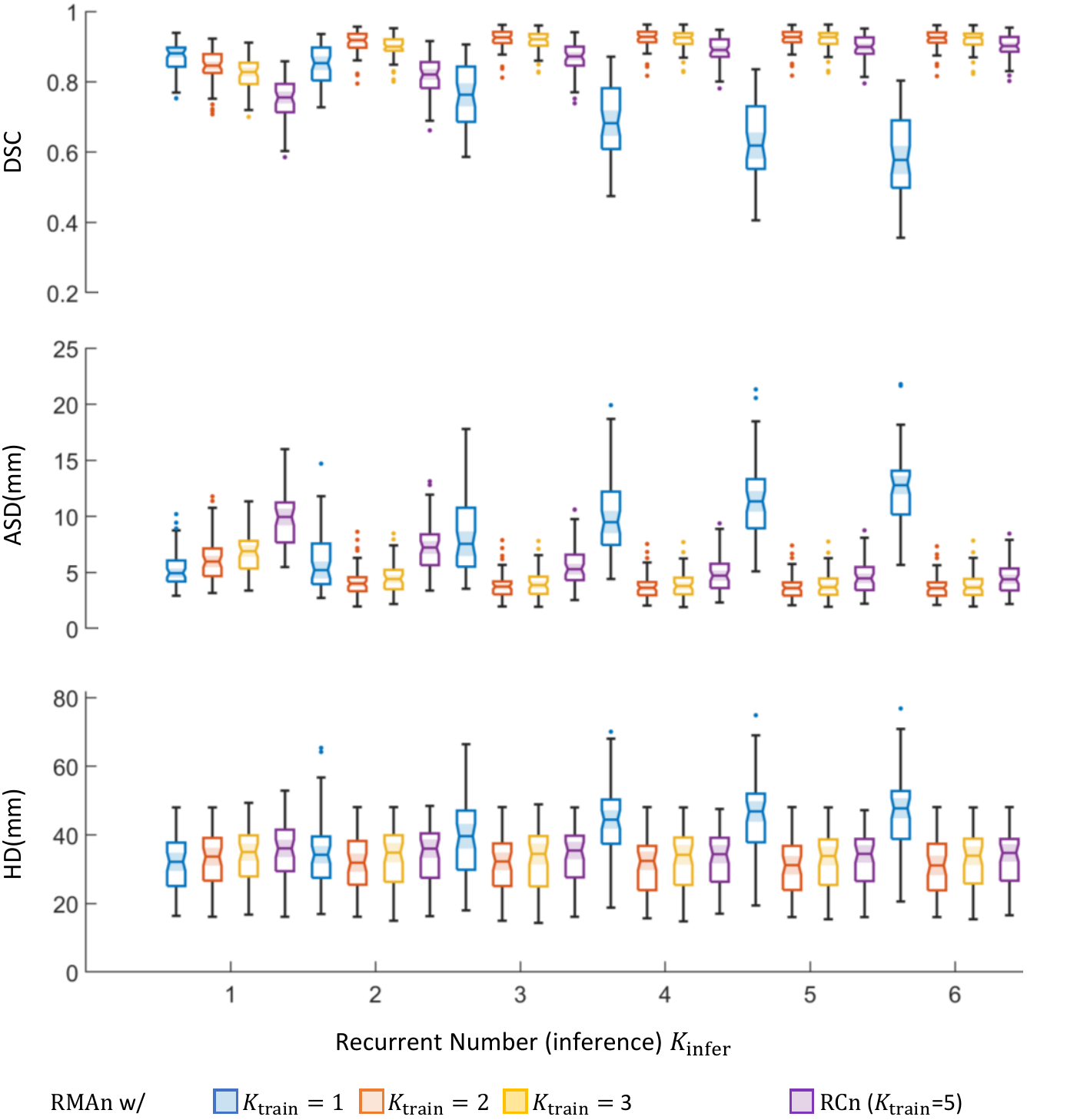}
\end{center}
\vspace{-1.5em}
\caption{The registration results on chest CT using our RMANs and the baseline RCn, with varying recursive number both for training and inference, shows that, with the increase of recursive number (inference), the model with recursive number (training) 2 and 3 achieve higher accuracy and converge closely, while it get worse with recursive number (training) 1, and RMAn outperform RCN with each $K_{\rm infer}$ in terms of DSC and ASD.}
\label{fig:quan_result_rec}
\end{figure}


\subsubsection{Ablation Study:}
Comparing VM1 and D-net in Tab.~\ref{tab:result_abdomen_lung}, the MA based architecture outperforms the pure encoder-decoder structure in two dataset with comparable network scale.
To validate the effect of recursive architecture, we also tried several combination on varying recursive number for training and testing stage respectively on experiments of abdomen and lung CT as shown in Tab.~\ref{tab:result_ablation} and Fig.~\ref{fig:quan_result_rec}. Comparing RMAn (${K_{\rm train}=1,K_{\rm infer}=1}$) with others, the results show recursive architecture used in both training and testing phase results in the improved accuracy both in chest and abdomen CT scans, and the larger recurrent number for training could bring significant improvement. In addition,  architecture reduces the negative number of Jacobian determinant, which thus improves the rationality of registration.

\subsubsection{Number of Recurrent Stages:}
\label{sec:results_rec}
Furthermore, RMAn is tested with varying recurrent number for both training and inference as shown in Fig.~\ref{fig:quan_result_rec}.
Surprisingly, the performance of RMAn with recurrent number $K_{\rm train}=1$ and $K_{\rm infer}>1$ for training and inference is even worse than MAn ($K_{\rm train}=1$ and $K_{\rm infer}=1$). This is probably due to the lack of recursive pattern during training for $K_{\rm train}=1$. As shown in Fig.~\ref{fig:quan_result_rec}, the RMAn with $K_{\rm train}=2$ and $K_{\rm train}=3$ as well as the RCn achieve improvement with more $K_{\rm infer}$. 
We also compare our RMAn with baseline RCn based on varying $K_{\rm infer}$ as shown in Fig.~\ref{fig:quan_result_rec}. It shows RMAn outperform RCn for varying $K_{\rm infer}\in[1,8]$ in terms of DSC and ASD.



\section{Discussion and Conclusion}
The novel RMAn design is proposed based on the MA structure incorporated in a recursive architecture. It achieves the best registration results in the inter-subject lung CT registration and one of the best on other 9 organs in abdominal CT scans compared with the state of the art networks. The recursive architectures for registration are also investigated via varying training and inference recurrent number. The results show that larger inference recurrent number can improve the registration results, and on the other hand, also implies a small influence of the training recurrent number as long as the sub-network is able to learn the pattern of recursive registration. The comparison of RMAn with RCn also proves the accuracy improvement stemming from the MA. In future, the proposed RMAn will be also applied to multi-modal image registration. 

\section*{Acknowledgements}
This work was supported by a Kennedy Trust for Rheumatology Research Studentship, the Centre for OA Pathogenesis Versus Arthritits (Versus Arthritis grant 21621). 
B. W. Papież acknowledges Rutherford Fund at Health Data Research UK (MR/S004092/1)
%
%
%
\bibliographystyle{splncs04}
\bibliography{ref}
%




\end{document}